\documentclass{article}

\usepackage[preprint]{neurips_2026}

\usepackage[utf8]{inputenc} % allow utf-8 input
\usepackage[T1]{fontenc}    % use 8-bit T1 fonts
\usepackage{hyperref}       % hyperlinks
\usepackage{url}            % simple URL typesetting
\usepackage{booktabs}       % professional-quality tables
\usepackage{amsfonts}       % blackboard math symbols
\usepackage{amsmath}
\usepackage{nicefrac}       % compact symbols for 1/2, etc.
\usepackage{microtype}      % microtypography
\usepackage{xcolor}
\usepackage{colortbl}   % \cellcolor for the value gradient
\usepackage{diagbox}     % \diagbox: diagonal split header cell
\usepackage{arydshln}    % \cdashline: dashed rule between supervision blocks
\definecolor{gradblue}{HTML}{3D7BC4}
\usepackage{xspace}
\definecolor{opsdpurple}{HTML}{7B3FA0}

\definecolor{cOurs}{HTML}{57BE93}   % green  -- our method (headline series)
\definecolor{cOursB}{HTML}{A3DCC1}  % light green -- secondary variant
\definecolor{cSup}{HTML}{74A6E3}    % blue   -- supervised baseline
\definecolor{cAlt}{HTML}{E8736C}    % red    -- supervised OPSD in Figure~\ref{fig:curves}
\definecolor{cBase}{HTML}{C6CBD2}   % grey   -- base model
\definecolor{cBlueS}{HTML}{6BAED6}  % blue  -- Figure~\ref{fig:config} series
\definecolor{cRedS}{HTML}{E8736C}   % red   -- Figure~\ref{fig:config} series
\definecolor{cGreyS}{HTML}{8C96A0}  % slate -- third series in Figure~\ref{fig:refgrid}
\definecolor{cInk}{HTML}{1E232A}    % ink    -- axes and labels

\usepackage{multirow}
\usepackage{algorithm}
\usepackage{algpseudocode}
\usepackage{pgfplots}
\usepackage{float}
\usepackage{amssymb}
\usepackage{bbm}
\pgfplotsset{compat=1.17}
\usepgfplotslibrary{groupplots}

\newcommand{\method}{\textsc{u-OPSD}\xspace}
\title{On-Policy Self-Distillation without Any Supervision}
\author{
Yijiang Li$^{\spadesuit}$ \quad
Bingyang Wang$^{\heartsuit}$ \quad
Yijun Liang$^{\diamondsuit}$ \quad
Yunjie Tian$^{\clubsuit}$ \quad
\\
\textbf{Di Fu}$^{\clubsuit}$ \quad
\textbf{Nuno Vasconcelos}$^{\spadesuit}$ \\[0.35em]
$^{\spadesuit}$UC San Diego \quad
$^{\heartsuit}$Georgia Institute of Technology \\
$^{\diamondsuit}$University of Maryland, College Park \quad
$^{\clubsuit}$ByteDance \\[0.3em]
\texttt{\{yijiangli, nuno\}@ucsd.edu
}
}

\begin{document}

\maketitle

\begin{abstract}
On-policy (Self-)Distillation (OPD / OPSD) has shown strong potential for post-training large language models (LLMs). However, existing methods still rely heavily on external supervision, including ground-truth signals, environmental feedback, or guidance from larger models, and therefore fall short of genuine ``self''-distillation. In this study, we show that on-policy self-distillation can be achieved using only a model's own generations via internal consistency.  
We propose unsupervised on-policy self-distillation (\method). \method first samples multiple rollouts and constructs a pseudo solution by majority vote under a self-consistency threshold. It then conditions the model's distribution on the pseudo-solution and distills itself on the disagreeing completions, allowing the model to correct itself precisely where it is confidently wrong.
Across diverse benchmarks, base models, and training settings, \method consistently improves over the base models and matches or surpasses supervised methods with ground truth (GT) such as OPSD and GRPO.
On five mathematical reasoning benchmarks, i.e., AIME24, AIME25, HMMT25, MATH500, and AMC23, \method improves over the base model by $8.5\%$ and $10.7\%$ on Qwen3 non-thinking mode at $4$B and $8$B scales, and outperforms OPSD by $3.2\%$ and $2.3\%$ on average, respectively. In thinking mode, \method stays on par with OPSD, ahead by $0.9\%$ at $4$B and level at $8$B and surpassing GRPO by $0.7\%$ and $1.1\%$, respectively. Code is available at  \url{https://github.com/williamium3000/u-opsd}.\footnote{Project Page: \url{https://williamium3000.github.io/u-opsd/}}
\end{abstract}

\section{Introduction}
\label{sec:intro}

Post-training has emerged as a key driver of advances in the reasoning capabilities of large language models (LLMs), with progress largely propelled by supervised fine-tuning (SFT) \citep{ye2025limo,wen2025lightr1,li2026less},
knowledge distillation from stronger teachers \citep{abdin2025phi4reasoning,lu2025onpolicydistillation},
and reinforcement learning with verifiable rewards (RLVR) \citep{minimax2025m1,guo2025deepseekr1,5team2025glm45agenticreasoningcoding}.

Among the many recipes, \emph{on-policy distillation} bridges SFT and RL by training on the model's own generations, reducing the train-inference mismatch~\citep{gu2024minillm,agarwal2024gkd} and catastrophic forgetting~\citep{shenfeld2026rls} of teacher-forced SFT, while retaining dense token-level supervision instead of sparse scalar rewards. Subsequent work has continued the effort of OPD along its objective, supervision, and systems dimensions. DistiLLM ~\citep{ko2024distillm} introduces skewed KL, while DistiLLM-2 applies asymmetric objectives to teacher- and student-generated responses~\citep{ko2025distillm2}. Other work stabilizes long-horizon OPD by restricting supervision to teacher-supported tokens and masking unreliable signals~\citep{fu2026revisiting}, or reduces its systems overhead by precomputing teacher scores under a teacher-consistent offline pipeline~\citep{wu2026lightning}. More recent analyses further identify teacher--student mismatch and length exploitation as key failure modes, motivating clipped and log-compressed token-level guidance~\citep{wang2026demystifying}. Nevertheless, these methods still heavily rely on a separate strong teacher model for guidance.

% need a section of progress in OPD here

On-policy \emph{self}-distillation (OPSD)~\citep{zhao2026self} pushes this further by removing the need for a stronger teacher model -- a single LLM plays both roles, where the teacher (student itself) conditions on the ground-truth solution, whereas the student sees only the problem. Subsequent work identifies information leakage and instability \citep{yang2026selfdistilled}, redesigns the privileged context \citep{ye2026opcd, penaloza2026privileged, sang2026crisp}, and extends it to agentic setups \citep{liu2026hero, yang2026selfdistilled}.

% need a section of progress in OPSD here

The privileged context, however, is also its own bottleneck. Existing OPD and OPSD methods rely on external supervision, such as ground-truth solutions, environmental feedback, or guidance from larger models. Thus, the model is self-distilled only in the sense that the teacher and student share parameters; the information that makes the teacher more capable still comes from outside the model. This dependence limits scalability to unlabeled problems and restricts applicability in domains where supervision is costly, unreliable, or unavailable.
We therefore ask: \emph{Does the teacher in on-policy self-distillation need a ground-truth solution? Can a model construct its own privileged context and perform genuine self-distillation?}

We answer with \method, an unsupervised on-policy self-distillation method that requires no external supervision. 
Our key observation is that although an individual rollout may be unreliable, agreement among multiple independently sampled rollouts provides an endogenous confidence signal. This makes it possible to derive both the teacher reference and the student trajectories entirely from the model's own on-policy samples.
Formally, for each unlabeled problem $x\sim\mathcal{U}$, we sample $G$ independent rollouts
$y^{(1)},\ldots,y^{(G)} \overset{\mathrm{i.i.d.}}{\sim} \bar\pi(\cdot\mid x)$, where $\mathcal{U}$ is the unlabeled problem distribution, $G$ is the number of rollouts per problem, and $\bar\pi\triangleq\pi_{\operatorname{sg}[\theta]}$ denotes a stop-gradient copy of the current policy $\pi_\theta$.
We extract the final answers from each rollout as $a^{(g)}=\mathrm{Ans}(y^{(g)})$, where $\mathrm{Ans}(\cdot)$ denotes the answer-extraction and canonicalization function, and obtain the vote-based \emph{pseudo-answer}
$\tilde{a}(x)=\operatorname*{arg\,max}_{a}\sum_{g=1}^{G}\mathbbm{1}\!\left[a^{(g)}=a\right]$, where $\mathbbm{1}[\cdot]$ is the indicator function.

The rollouts are then partitioned into agreeing and disagreeing multisets $\mathcal{Y}^{+}_{x}=\{y^{(g)}:a^{(g)}=\tilde{a}(x)\}, \qquad \mathcal{Y}^{-}_{x} = \{y^{(g)}:a^{(g)}\neq\tilde{a}(x)\}$.
When the winning vote fraction reaches a self-consistency threshold $\tau$, the longest agreeing rollout $y^{+} \in \operatorname*{arg\,max}_{y\in \mathcal{Y}^{+}_{x}} |y|$ serves as the \emph{pseudo-solution}, while the disagreeing rollouts in $\mathcal{Y}^{-}_{x}$ serve as
the student trajectories. By distilling the teacher distribution conditioned on the pseudo-solution into the student along prefixes of the model's disagreeing completions, \method achieves continual improvement through on-policy self-distillation without any external supervision. 
\begin{equation}
\begin{aligned}
\mathcal{L}_{\method}(\theta)
= \mathbb{E}_{x\sim\mathcal{U}}\,
\mathbb{E}_{\{y^{(g)}\}_{g=1}^{G}\sim \bar\pi(\cdot\mid x)}
&\Bigg[
\mathbbm{1}\!\left[\mathcal{Y}^{-}_{x}\neq\emptyset\right]
\frac{1}{|\mathcal{Y}^{-}_{x}|}
\sum_{y^{-}\in\mathcal{Y}^{-}_{x}}
\frac{1}{|y^{-}|}
\sum_{n=1}^{|y^{-}|}
\\
&\qquad
D_{\beta}\!\left(
\bar\pi\!\left(\cdot\mid x,\,y^{+},\,y^{-}_{<n}\right)
\,\middle\|\,
\pi_\theta\!\left(\cdot\mid x,\,y^{-}_{<n}\right)
\right)
\Bigg].
\end{aligned}
\end{equation}
Here, $|y^{-}|$ denotes the token length of a disagreeing rollout, and
$y^{-}_{<n}$ denotes its prefix before token position $n$.
The function $D_{\beta}(P\,\|\,Q)$ denotes a divergence between the teacher and student next-token distributions, such as the forward KL divergence
$D_{\mathrm{KL}}(P\,\|\,Q)$.

We evaluate \method on five mathematical reasoning benchmarks across six Qwen3 configurations: 4B and 8B models in both non-thinking and thinking modes, together with Qwen3-4B-Instruct-2507 and Qwen3-30B-A3B-Instruct-2507. Across all configurations, \method improves over the base model, by $8.5$--$10.7\%$ in non-thinking mode, $1.9$--$2.2\%$ in thinking mode, and $1.7$--$1.8\%$ on the Instruct models. \method also improves over prior self-rewarding RL methods by a large margin, i.e., on average $7.0$--$11.3\%$ in non-thinking mode and $0.8$--$1.4\%$ in thinking mode, showing that consensus is substantially more effective as conditioning context for token-level distillation than as a scalar reward for policy optimization. Without any external supervision, \method is still on par with or better than supervised SFT, GRPO, and OPSD with GT labels, with gains of up to $10.9\%$, $8.9\%$, and $3.2\%$, respectively, while tying OPSD only in the 8B thinking setting.

\section{Related work}
\label{sec:related}
% \textcolor{red}{survey, exclude recent month}

\textbf{Self-rewarding Reinforcement Learning.}
Although RLVR has proven effective at strengthening LLM reasoning \citep{shao2024deepseekmath, guo2025deepseekr1}, it hinges on curated ground-truth labels, whose cost and scarcity quickly become the limiting factor \citep{yue2025rlvr}. A growing body of work replaces external verification with reward signals the model derives from its own behavior on unlabeled data. Among these intrinsic signals, confidence-based rewards are especially appealing: a model's certainty in its own output---captured directly as self-certainty \citep{intuitor2025, li2025rlsc} or inversely through predictive entropy \citep{rent2025, zhang2025empo}---yields a dense, label-free reward that requires no extra sampling infrastructure and correlates well with answer correctness. Agreement across sampled solutions offers a complementary signal, as exploited by majority voting \citep{wang2023selfconsistency} and test-time training on self-consistency \citep{zuo2025ttrl}. These methods build on earlier self-rewarding language models \citep{yuan2024selfrewarding} and self-play supervision \citep{chen2024spin}, and the paradigm has since broadened to unsupervised self-training \citep{xu2025genius, fang2025serl}, self-correction guided by the model's own judgments \citep{xiong2025selfrewardingcorrection}, and zero-data self-evolution in which models construct their own curricula \citep{zhao2025absolutezero, huang2025rzero, liu2025spiral}.

\textbf{On-policy (self-) distillation.}
% \textbf{On-policy distillation and on-policy self-distillation.}
On-policy distillation trains on student-generated trajectories, allowing a teacher to provide dense token-level supervision at states visited by the student and reducing the train--inference mismatch of off-policy distillation \citep{gu2024minillm,agarwal2024gkd}.
On-policy self-distillation removes the separate teacher by letting the same model act under asymmetric contexts, with the teacher conditioned on verified solutions \citep{zhao2026self}, demonstrations or in-context examples \citep{shenfeld2026sdft}, privileged observations \citep{penaloza2026privileged}, or environmental feedback \citep{hubotter2026sdpo}.
Subsequent work explores alternative conditioning signals \citep{sang2026crisp}, derives process supervision from externally verified successful and unsuccessful trajectories \citep{tan2026ssopd}, or modulates token-level supervision according to teacher reliability \citep{ke2026uncertainty,liu2026pwopsd}.
\method instead constructs the teacher reference from rollout agreement and identifies correction targets through rollout disagreement, without annotations, demonstrations, environmental feedback, gold answers, or externally verified outcomes.

\textbf{Self-training, self-distillation and consistency.}
Self-training improves models using supervision derived from their own generations.
Existing methods select or refine self-generated data using answer verification and reward signals \citep{zelikman2022star,yuan2023scaling,gulcehre2023rest}, as well as confidence, consistency, and iterative self-improvement \citep{huang2023selfimprove,liang2026selfevolving}.
More broadly, self-distillation transfers knowledge across different instances or views of the same model, including previous model generations and intermediate optimization snapshots \citep{furlanello2018bornagain,yang2018snapshot}.
Other approaches distill across architectural branches or augmented views \citep{zhang2019beyourownteacher, li2023diverse, caron2021dino}, while recent work extends self-distillation to foundation models using diverse reasoning traces or multiple self-teachers \citep{wu2025sdrt,jin2026unisd}.
Self-consistency originally aggregates multiple reasoning paths through majority voting at inference time \citep{wang2023selfconsistency}, while subsequent methods turn agreement into sequence-level preferences \citep{prasad2024scpo}, pseudo-labels or scalar reinforcement-learning rewards \citep{zuo2025ttrl,shafayat2025selftrain}, or model-generated feedback and preferences \citep{yuan2024selfrewarding}.
\method connects self-consistency with on-policy self-distillation by using consensus to construct an instance-specific self-teacher and transferring dense next-token distributional supervision along prefixes of consensus-disagreeing rollouts.

\section{Method}
\label{method}

\subsection{Preliminaries}
\label{sec:prelim}

\begin{figure}[h]
    \centering
    \includegraphics[width=1\linewidth]{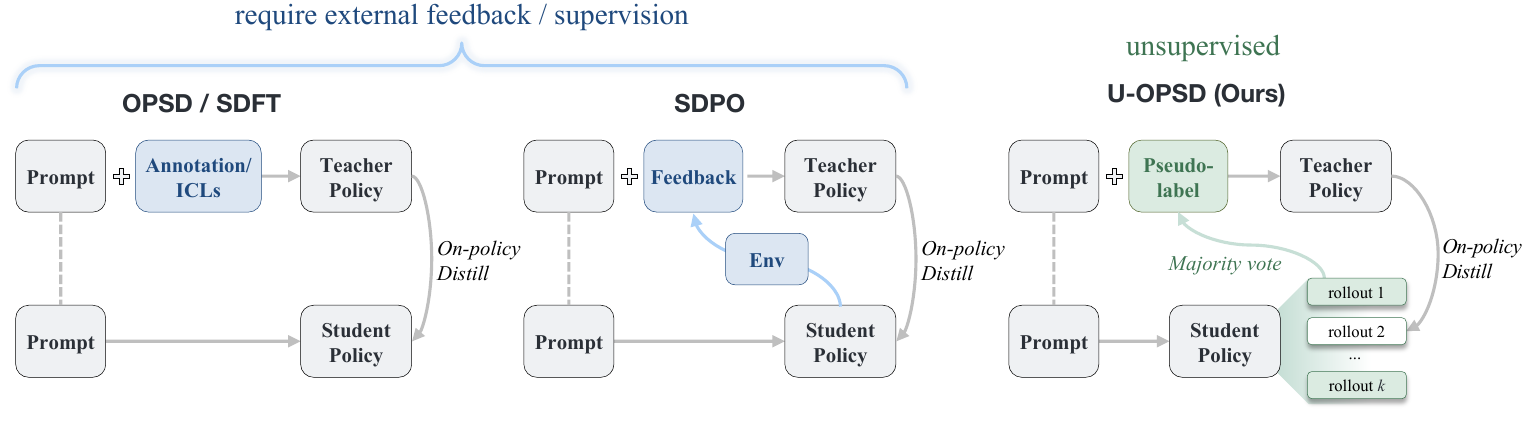}
    \caption{Comparison between OPSD / SDFT with ground-truth solution or ICLs (left), SDPO with rich feedback from the environment (middle) and our \method without any supervision. }
    \label{fig:comparison}
\end{figure}

\paragraph{Notation.}
Let $\pi_\theta$ be an LLM defining next-token distributions $\pi_\theta(\cdot\mid c)$
over a vocabulary $\mathcal{V}$ for a context $c$, and let
$\bar\pi\triangleq\pi_{\operatorname{sg}[\theta]}$ be the same network with gradients
detached, so that sampling from or scoring under $\bar\pi$ carries no gradient.
For a sequence $y$, $y_{<t}$ denotes its prefix, and
$\mathrm{Ans}(y)\in\mathcal{A}\cup\{\varnothing\}$ the final answer parsed from it,
where $\mathcal{A}$ is the space of admissible answers and $\varnothing$ marks a
generation from which no answer can be parsed.
Supervised post-training draws from a labeled corpus
$\mathcal{S}=\{(x,y^{\star})\}$ of problems with reference solutions and gold
answers $a^{\star}=\mathrm{Ans}(y^{\star})$; \method uses only unlabeled problems
$x\sim\mathcal{U}$. Throughout, $D_{\beta}$ denotes the generalized
Jensen--Shannon divergence between distributions on $\mathcal{V}$, whose
$\beta\to 0$ limit is the forward KL.

\paragraph{Group Relative Policy Optimization: sparse signal by verifiable reward.}
GRPO~\citep{shao2024deepseekmath} optimizes the policy with a verifiable reward
normalized within a group of sampled rollouts: for each problem it samples $G$
rollouts $y^{(1)},\ldots,y^{(G)}$, scores them with
$r^{(g)}=\mathbbm{1}\!\left[\mathrm{Ans}(y^{(g)})=a^{\star}\right]$, and normalizes
within the group,
$A^{(g)}=\big(r^{(g)}-\operatorname{mean}\{r^{(j)}\}_{j=1}^{G}\big)\big/
\operatorname{std}\{r^{(j)}\}_{j=1}^{G}$, so that all tokens of a rollout share one
sequence-level advantage. The policy is then updated on the clipped surrogate

\begin{equation}
\label{eq:grpo}
\begin{aligned}
\mathcal{L}_{\mathrm{GRPO}}(\theta)
=-\,\mathbb{E}_{x\sim\mathcal{S}}\,
\mathbb{E}_{\{y^{(g)}\}_{g=1}^{G}\sim\bar\pi(\cdot\mid x)}\,
\frac{1}{G}\sum_{g=1}^{G}\frac{1}{|y^{(g)}|}\sum_{t=1}^{|y^{(g)}|}
\min\!\Big(&\rho^{(g)}_{t}A^{(g)},\\[-2pt]
&\operatorname{clip}\big(\rho^{(g)}_{t},\,1{-}\varepsilon,\,1{+}\varepsilon\big)A^{(g)}\Big),
\end{aligned}
\end{equation}

where $\rho^{(g)}_{t}=\pi_\theta\big(y^{(g)}_{t}\mid x,y^{(g)}_{<t}\big)\big/
\bar\pi\big(y^{(g)}_{t}\mid x,y^{(g)}_{<t}\big)$ is the importance ratio to the
behaviour policy and $\varepsilon$ is the clipping range. The supervision is the gold answer $a^{\star}$, and the
signal is sparse: it is sequence-level, and vanishes whenever all $G$ rewards
coincide.

\paragraph{On-policy distillation: dense signal by external teacher.}
On-policy distillation (OPD) addresses this sparsity by retaining the model's own
rollouts as training trajectories while replacing the scalar reward with dense
supervision from a teacher's next-token distribution. The student is conditioned
only on the problem $x$, matching inference time, while the teacher receives a
possibly richer context $c$. Writing $\pi_{T}$ for the teacher, these methods share
the form

\begin{equation}
\label{eq:opd}
\mathcal{L}_{\mathrm{OPD}}(\theta)
=\mathbb{E}_{x}\,\mathbb{E}_{y\sim\bar\pi(\cdot\mid x)}
\sum_{t=1}^{|y|}
D_{\beta}\big(\pi_{T}(\cdot\mid c,\,y_{<t})\,\big\|\,\pi_\theta(\cdot\mid x,\,y_{<t})\big),
\end{equation}

which yields dense token-level supervision at states the student actually visits.
Variants differ only in the choice of teacher $\pi_{T}$ and the context $c$ it
receives. OPD~\citep{gu2024minillm,agarwal2024gkd,lu2025onpolicydistillation} takes
$\pi_{T}$ to be a separate, typically stronger model with $c=x$. The supervision is
dense, but it resides in an external teacher that must already surpass the
model being trained.

\paragraph{On-policy self-distillation.}
OPSD~\citep{zhao2026self} takes a further step by constructing the teacher and
student from the same language model under different conditioning contexts,
removing the external teacher and supplying a \emph{gold solution} in its
place. The student policy observes only the problem statement $x$, matching
inference time, while the teacher policy conditions on both the problem statement $x$ and the GT solution
$y^{\star}$:
\begin{equation*}
\pi_{S}(\cdot \mid x) \triangleq \pi_\theta(\cdot \mid x),
\qquad\qquad
\pi_{T}(\cdot \mid x, y^{\star}) \triangleq \bar\pi(\cdot \mid x, y^{\star}).
\end{equation*}
Given a problem $x$ paired with a GT solution $y^{\star}$, OPSD samples
the full next-token distribution from both policies, then minimizes a token-level
divergence between them:
\begin{equation}
\label{eq:opsd}
\mathcal{L}_{\mathrm{OPSD}}(\theta)
=\mathbb{E}_{(x,y^{\star})\sim\mathcal{S}}\,
\mathbb{E}_{y\sim\bar\pi(\cdot\mid x)}
\sum_{t=1}^{|y|}
D_{\beta}\big(\pi_{T}(\cdot\mid x,\,y^{\star},\,y_{<t})\,\big\|\,\pi_{S}(\cdot\mid x,\,y_{<t})\big).
\end{equation}
Because the teacher sees $y^{\star}$, its next-token distribution along the
student's trajectory encodes what a model that already knows the solution would do
at each position, and matching it distills solution-conditioned behavior into the
solution-free policy. Following the strongest OPSD configuration, we use
$\beta = 0$ (forward KL, teacher $\rightarrow$ student) with per-token point-wise
clipping for stability. Teacher and student share parameters, yet OPSD still requires external supervision, i.e., the reference solution $y^{\star}$.

\subsection{\method: On-Policy Self-Distillation without Any Supervision}
\label{sec:method:unsup}

\vspace{-2mm}

\begin{figure}[h]
    \centering
    \includegraphics[width=1\linewidth]{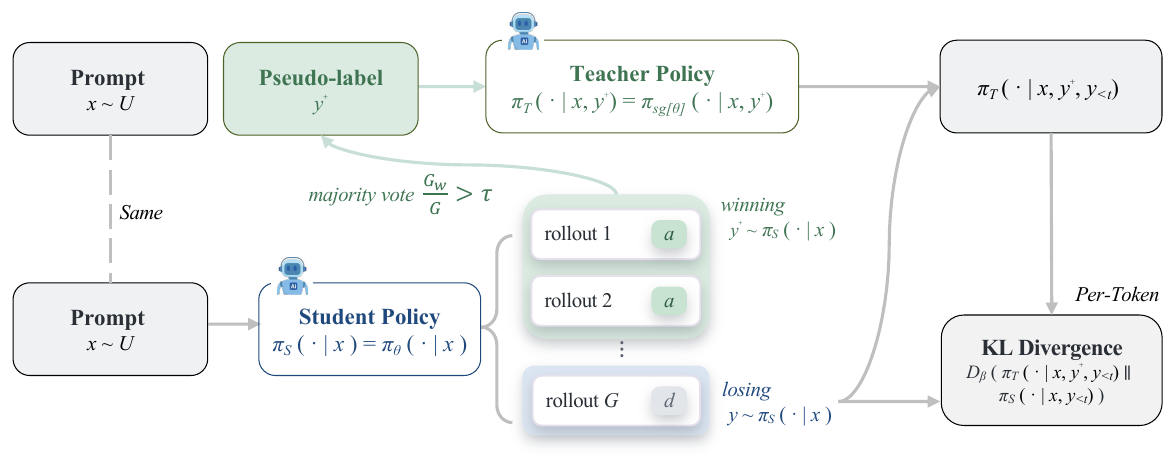}
    \caption{Overview of Unsupervised On-policy Self-Distillation (\method), which replaces ground-truth supervision in On-Policy Self-distillation with pseudo-labels generated from the model's own majority-vote consensus.}
    \label{fig:overview}
\end{figure}

Despite on-policy training, across Eqs.~\eqref{eq:opd}--\eqref{eq:opsd}, these methods remain externally supervised. RLVR relies on gold answers for verification, OPD on a stronger teacher, and OPSD and related work on privileged teacher contexts such as GT solutions~\citep{zhao2026self}, demonstrations~\citep{shenfeld2026sdft}, or environmental feedback~\citep{hubotter2026sdpo} (see the left of Figure~\ref{fig:comparison}). 
% Their ``self'' nature is therefore limited to parameter sharing: the information that makes the teacher more capable still originates outside the model. . This dependence restricts their applicability in domains where supervision is costly, unreliable, or unavailable, and ultimately precludes genuine model self-evolving.
\method removes this dependence by enabling the model to construct its own privileged context through internal consistency, i.e., a majority vote over its
own rollouts that identifies a pseudo-solution in place of
$y^{\star}$, while the resulting conditional teacher distribution is distilled into the student on the disagreeing rollouts (Figure~\ref{fig:comparison}, right).

\textbf{(1) Sample.} For each prompt, draw $G$ rollouts $y^{(1)}, \dots, y^{(G)} \sim \bar\pi(\cdot \mid x)$ at the training temperature and parse
$a^{(g)}=\mathrm{Ans}(y^{(g)})\in\mathcal{A}\cup\{\varnothing\}$ (i.e., extract the final answer $a^{(g)}$ from \verb|\boxed{...}| with normalization and canonicalization). 
$\varnothing$ marks rollouts without a parsable final answer (usually truncations), which we call \emph{invalid}.

\textbf{(2) Vote.} Among the valid answers, we define the pseudo-answer $\tilde{a}(x)$ as the majority vote, breaking ties uniformly at random~\citep{wang2023selfconsistency}. This vote partitions the rollouts into agreeing and disagreeing
sets $\mathcal{Y}^{+}_{x}=\{y^{(g)}:a^{(g)}=\tilde{a}(x)\}, \qquad \mathcal{Y}^{-}_{x} = \{y^{(g)}:a^{(g)}\neq\tilde{a}(x)\}$. Invalid rollouts belong to neither set, as an incomplete generation provides evidence of neither a correct nor an incorrect belief.
We quantify vote confidence using the self-consistency score
$
c(x)=\frac{1}{G}\sum_{g=1}^{G}\mathbbm{1}\!\left[a^{(g)}=\tilde{a}(x)\right].
$
The score is normalized by the total number of sampled rollouts $G$, rather than only by the number of valid rollouts where truncated generations reduce confidence. If $c(x)<\tau$, the prompt is treated as unlabeled and contributes no gradient in the training step. We set $\tau=\nicefrac{1}{2}$ for an absolute majority.

\textbf{(3) Distill.}
Given a valid majority vote, we select an agreeing rollout $y^{+}\in\mathcal{Y}^{+}_{x}$ as the teacher reference in place of $y^{\star}$, and a subset $\mathcal{B}^{-}_{x}\subseteq\mathcal{Y}^{-}_{x}$ of disagreeing rollouts as the distillation targets. Substituting the pseudo solution $(x,y^{+})$ into Eq.~\eqref{eq:opsd} yields

\begin{equation}
\label{eq:unsup}
\mathcal{L}_{\method}(\theta) \;=\; \frac{1}{|\mathcal{B}^{-}_{x}|}\sum_{y^{-} \in \mathcal{B}^{-}_{x}} \sum_{t=1}^{|y^{-}|} D_{\beta}\!\Big( \bar\pi\big(\cdot \mid x, y^{+}, y^{-}_{<t}\big) \;\Big\|\; \pi_\theta\big(\cdot \mid x, y^{-}_{<t}\big) \Big).
\end{equation}

 % We deliberately leave the selection of reference unspecified in the general objective, as they define a design space with no uniformly dominant choice; we examine several alternatives under \emph{Design variants} below. When $\mathcal{B}^{-}{x}=\mathcal{Y}^{-}_{x}$ and $\tau\to 0$, Eq.~\eqref{eq} recovers the idealized objective in Eq.~\eqref{eq}, up to the exclusion of unparsable rollouts.

% Following the strongest OPSD configuration, we instantiate $\mathcal{D}$ as forward KL, corresponding to the normalized $\beta\to 0$ limit of $\mathrm{JSD}{\beta}$ under the convention $\mathrm{JSD}{0}\triangleq D_{\mathrm{KL}}(P,|,Q)$, and apply per-token clipping for stability (Section~\ref{sec:exp}).

The above skips two classes of samples: those for which the model cannot form a sufficiently consistent vote ($c(x)<\tau$), making the pseudo-answer unreliable, and those for which all valid rollouts already agree ($\mathcal{Y}^{-}_{x}=\emptyset$), leaving no disagreement to correct. Training therefore focuses on the model's competence frontier: prompts for which it can consistently identify a plausible solution but still assigns substantial probability to conflicting trajectories. This curriculum emerges directly from the model's own voting statistics, without external difficulty labels or scheduling.

Unlike OPSD, \method requires no labeled pair $(x,y^{\star})$; its privileged context is constructed entirely from the model's own on-policy rollouts. Unlike conventional majority-vote self-training, it does not simply imitate the selected response under teacher-forced prefixes. Instead, it transfers the solution-conditioned next-token distribution along the model's own disagreeing trajectories, providing dense corrective supervision precisely at the prefixes that lead toward answers inconsistent with the model's consensus. In this way, \method enables iterative on-policy self-evolution and self-improvement.

% \textbf{(3) Select the teacher reference and the distillation targets.} Promote the \emph{shortest} agreeing rollout, $\hat s = \arg\min_{i:\, a^{(i)} = \hat a} |y^{(i)}|$, to the teacher's reference solution: among the self-consistent rollouts, the shortest tends to be the cleanest derivation and keeps the teacher context compact. Collect the \emph{confidently wrong} set $W = \{ i \in V : a^{(i)} \neq \hat a \}$: valid rollouts whose final answer contradicts the consensus. Invalid rollouts are excluded from $W$---an unfinished generation is not evidence of a wrong belief, and we found distilling such rollouts unhelpful. Select at most $k$ elements of $W$ (the \emph{row cap}; our default is $k=1$, selected uniformly at random) as distillation targets $W_k$.

% \textbf{(4) Distill.} Apply the OPSD objective of Eq.~\eqref{eq:opsd} with $s$ replaced by $\hat s$, summing over the selected disagreeing rollouts only:
% \begin{equation}
% \label{eq:unsup}
% \mathcal{L}_{\method}(\theta) \;=\; \sum_{i \in W_k} \sum_{t=1}^{|y^{(i)}|} D_{\beta}\!\Big( \bar\pi\big(\cdot \mid x, \hat s, y^{(i)}_{<t}\big) \;\Big\|\; \pi_\theta\big(\cdot \mid x, y^{(i)}_{<t}\big) \Big).
% \end{equation}

% Prompts where all valid rollouts agree also contribute no gradient: the model is already self-consistent there, and the teacher would have nothing to correct. Training therefore concentrates its entire budget on the frontier where the model can produce the (self-believed) right answer yet still frequently produces wrong ones---a curriculum that emerges for free from the voting statistics.
\begin{algorithm}[h]
\caption{\method: unsupervised on-policy self-distillation (one prompt)}
\label{alg:unsup}
\begin{algorithmic}[1]
\Require prompt $x$; student $\pi_\theta$; detached teacher $\bar\pi$; rollouts $G$; threshold $\tau$
\State sample $y^{(1)},\ldots,y^{(G)}\overset{\mathrm{i.i.d.}}{\sim}\bar\pi(\cdot\mid x)$ \Comment{on-policy, training temperature}
\State $a^{(g)}\gets\mathrm{Ans}(y^{(g)})$ for all $g$ \Comment{$\varnothing$ if unparsable}
\State $\tilde{a}(x)\gets$ plurality of $\{a^{(g)}:a^{(g)}\neq\varnothing\}$;\quad $c(x)\gets\frac{1}{G}\sum_{g}\mathbbm{1}[a^{(g)}=\tilde{a}(x)]$
\State $\mathcal{Y}^{+}_{x}\gets\{y^{(g)}:a^{(g)}=\tilde{a}(x)\}$;\quad $\mathcal{Y}^{-}_{x}\gets\{y^{(g)}:a^{(g)}\notin\{\tilde{a}(x),\varnothing\}\}$
\If{$c(x)<\tau$ \textbf{or} $\mathcal{Y}^{-}_{x}=\emptyset$}
\State \Return \Comment{vote not trusted, or nothing to correct}
\EndIf
\State select $y^{+}\in\mathcal{Y}^{+}_{x}$ and $\mathcal{B}^{-}_{x}\subseteq\mathcal{Y}^{-}_{x}$
\State minimize $\displaystyle\frac{1}{|\mathcal{B}^{-}_{x}|}\sum_{y^{-}\in\mathcal{B}^{-}_{x}}\ \sum_{t=1}^{|y^{-}|} D_{\beta}\big(\bar\pi(\cdot\mid x,y^{+},y^{-}_{<t})\,\|\,\pi_\theta(\cdot\mid x,y^{-}_{<t})\big)$
\end{algorithmic}
\end{algorithm}

% Unlike standard OPSD, \method requires no labeled pair $(x,y^{\star})$: its privileged context is constructed entirely from the model's own on-policy rollouts. Unlike conventional majority-vote self-training, it does not simply imitate the selected response under teacher-forced prefixes. Instead, it transfers dense token-level guidance from an agreeing pseudo-solution along the model's own disagreeing trajectories. In this way, \method enables iterative on-policy improvement using only unlabeled problems and the model itself.

% \paragraph{Design variants.} The teacher reference is always the shortest agreeing rollout; the variants we ablate in Section~\ref{sec:exp:ablations} differ only in how the distillation targets are drawn from $W$: our main method \emph{disagree-1} takes $k{=}1$ uniformly at random; \emph{disagree-2} takes $k{=}2$ at random; and \emph{longest-1} takes the single \emph{longest} disagreeing rollout, on the intuition that long wrong rollouts contain the most tokens to correct.

\section{Experiments}
\label{sec:exp}

\subsection{Setup}
\label{sec:exp:setup}

\textbf{Models and datasets.}
We adopt four variants of Qwen3~\citep{yang2025qwen3}, including Qwen3-4B, Qwen3-8B, Qwen3-30B-A3B-Instruct-2507 and Qwen3-4B-Instruct-2507. For Qwen3-4B and Qwen3-8B, we train them in both non-thinking and thinking modes. In all experiments, the same checkpoint serves as both student and teacher. For the training set, we adopt a 30k subset of OpenThoughts~\citep{guha2025openthoughts} following OPSD; \method uses only the problem statements, never the solution field.
We compare against three supervised baselines trained on the same prompts: SFT on their gold solutions, GRPO with binary outcome rewards verified against the gold answer, and OPSD, which conditions the teacher on the gold solution. All OPSD numbers we report are from our own rerun with the released code and hyperparameters, under conditions identical to the \method runs. 

\textbf{Evaluation and benchmarks.}
We adopt five competition-level math reasoning benchmarks: AIME24, AIME25, and HMMT25 at avg@12, and MATH500~\citep{hendrycks2021math,lightman2024lets} and AMC23 at avg@4. We follow OPSD's protocol for decoding using vLLM at temperature 1.0 with maximum generation length 38k, in the same reasoning mode the model was trained in. 

Following OPSD's setup, we evaluate checkpoints every 25 steps up to 150 steps and report the best score for \method and OPSD, For GRPO, we report the peak performance within 500 training steps. For SFT, we train on the same number of samples as OPSD. 
For TTRL, RENT and Intuitor, we adopt the same training setup as \method, i.e., training with the same rollout budget and the same prompt set.

\textbf{Implementation details.}
We follow most of OPSD's training recipe~\citep{zhao2026self}: we use forward KL ($\beta{=}0$) over the full vocabulary with per-token pointwise clipping as the objective. We also adopt a teacher fixed to the initial policy rather than the running one to ensure fair comparison. Both choices are supported by ablations in \citet{zhao2026self}. We use LoRA of rank 64 ($\alpha{=}128$) on all attention and MLP projections, a learning rate of $5{\times}10^{-6}$, gradient-norm clipping at 0.1, and sampling at temperature 1.1 with top-$p$ 0.95 and top-$k$ 20. All runs are trained for 150 steps with a checkpoint every 25.

 For \method, we introduce two unique hyperparameters: for each prompt, we generate $G{=}8$ rollouts independently, filtered by confidence threshold $\tau{=}0.5$ (default unless specified otherwise), where supervised OPSD instead draws a single rollout from each of 32 prompts per optimizer step. We also increase the maximum completion length from 1,024 to 4,096 tokens, as \method requires rollouts to reach a boxed final answer; \citet{zhao2026self} report comparable performance between these two token budgets in their setting.
 For thinking-mode experiments, following OPSD, we keep the teacher in thinking mode and distill its behavior into a student in non-thinking mode, while evaluating the resulting model in thinking mode. For non-thinking experiments, both the teacher and student are in non-thinking mode during training, and evaluation is likewise conducted in non-thinking mode.

\begin{table}[h]
\caption{Performance comparison on math reasoning benchmarks for Qwen3 models with non-thinking mode. }
\label{tab:scale_nt}
\centering
\small
\setlength{\tabcolsep}{5pt}
\begin{tabular}{cl|cccccc}
\toprule
 & \textbf{Method} & \textbf{AIME24} & \textbf{AIME25} & \textbf{HMMT25} & \textbf{MATH500} & \textbf{AMC23} & \textbf{Avg.} \\
\midrule
\multicolumn{8}{l}{\textit{Qwen3-4B}} \\
 & \quad Base & \cellcolor{gradblue!8}25.83 & 17.78 & 10.83 & \cellcolor{gradblue!6}84.10 & 66.25 & 40.96 \\
\multirow{3}{*}{\emph{w/ GT.}} & \quad + SFT & \cellcolor{gradblue!9}26.67 & \cellcolor{gradblue!6}19.72 & \cellcolor{gradblue!13}13.06 & \cellcolor{gradblue!14}84.85 & \cellcolor{gradblue!7}69.38 & \cellcolor{gradblue!7}42.73 \\
 & \quad + GRPO & \cellcolor{gradblue!6}25.00 & \cellcolor{gradblue!15}22.50 & \cellcolor{gradblue!24}15.00 & \cellcolor{gradblue!29}86.20 & \cellcolor{gradblue!31}80.62 & \cellcolor{gradblue!18}45.86 \\
 & \quad + OPSD & \cellcolor{gradblue!21}32.22 & \cellcolor{gradblue!10}20.83 & \cellcolor{gradblue!32}\textbf{16.39} & \cellcolor{gradblue!24}85.75 & \cellcolor{gradblue!21}76.25 & \cellcolor{gradblue!20}46.29 \\
\addlinespace[1.5pt]
\cdashline{2-8}[1.2pt/1.4pt]
\addlinespace[1.5pt]
\multirow{4}{*}{\emph{w/o GT.}} & \quad + TTRL & \cellcolor{gradblue!6}25.00 & \cellcolor{gradblue!10}20.83 & \cellcolor{gradblue!6}11.94 & 83.60 & \cellcolor{gradblue!3}67.50 & \cellcolor{gradblue!3}41.77 \\
 & \quad + RENT & 22.22 & \cellcolor{gradblue!8}20.28 & \cellcolor{gradblue!3}11.39 & \cellcolor{gradblue!4}84.00 & \cellcolor{gradblue!17}74.38 & \cellcolor{gradblue!6}42.45 \\
 & \quad + Intuitor & \cellcolor{gradblue!3}23.89 & \cellcolor{gradblue!7}20.00 & \cellcolor{gradblue!5}11.67 & \cellcolor{gradblue!1}83.70 & \cellcolor{gradblue!9}70.62 & \cellcolor{gradblue!4}41.98 \\
 & \quad + \method & \cellcolor{gradblue!32}\textbf{37.50} & \cellcolor{gradblue!32}\textbf{27.78} & \cellcolor{gradblue!21}14.44 & \cellcolor{gradblue!32}\textbf{86.50} & \cellcolor{gradblue!32}\textbf{81.25} & \cellcolor{gradblue!32}\textbf{49.49} \\
\midrule
\multicolumn{8}{l}{\textit{Qwen3-8B}} \\
 & \quad Base & \cellcolor{gradblue!2}27.50 & \cellcolor{gradblue!5}23.33 & \cellcolor{gradblue!11}13.61 & 84.05 & 69.38 & \cellcolor{gradblue!2}43.57 \\
\multirow{3}{*}{\emph{w/ GT.}} & \quad + SFT & \cellcolor{gradblue!1}26.94 & \cellcolor{gradblue!1}21.67 & \cellcolor{gradblue!5}11.94 & \cellcolor{gradblue!1}84.10 & \cellcolor{gradblue!6}72.50 & \cellcolor{gradblue!1}43.43 \\
 & \quad + GRPO & \cellcolor{gradblue!7}30.56 & \cellcolor{gradblue!2}21.94 & \cellcolor{gradblue!9}13.06 & \cellcolor{gradblue!22}87.85 & \cellcolor{gradblue!9}73.75 & \cellcolor{gradblue!7}45.43 \\
 & \quad + OPSD & \cellcolor{gradblue!26}41.67 & \cellcolor{gradblue!16}28.06 & \cellcolor{gradblue!31}18.33 & \cellcolor{gradblue!18}87.15 & \cellcolor{gradblue!32}\textbf{85.00} & \cellcolor{gradblue!26}52.04 \\
\addlinespace[1.5pt]
\cdashline{2-8}[1.2pt/1.4pt]
\addlinespace[1.5pt]
\multirow{4}{*}{\emph{w/o GT.}} & \quad + TTRL & \cellcolor{gradblue!2}27.22 & 21.11 & \cellcolor{gradblue!9}13.06 & \cellcolor{gradblue!3}84.45 & \cellcolor{gradblue!5}71.88 & \cellcolor{gradblue!2}43.54 \\
 & \quad + RENT & \cellcolor{gradblue!4}28.33 & \cellcolor{gradblue!1}21.67 & 10.83 & 84.00 & \cellcolor{gradblue!1}70.00 & 42.97 \\
 & \quad + Intuitor & 26.11 & \cellcolor{gradblue!3}22.50 & \cellcolor{gradblue!3}11.67 & \cellcolor{gradblue!1}84.20 & \cellcolor{gradblue!8}73.12 & \cellcolor{gradblue!2}43.52 \\
 & \quad + \method & \cellcolor{gradblue!32}\textbf{45.56} & \cellcolor{gradblue!32}\textbf{34.72} & \cellcolor{gradblue!32}\textbf{18.61} & \cellcolor{gradblue!32}\textbf{89.55} & \cellcolor{gradblue!28}83.12 & \cellcolor{gradblue!32}\textbf{54.31} \\
\bottomrule
\end{tabular}
\end{table}

\subsection{Main results}
\label{sec:exp:main}

% [commented out on request] \paragraph{Cost of the consensus signal.} Removing labels is not free, because the pseudo-solution has to be paid for in generation. Supervised OPSD draws a single 1024-token rollout per prompt and distills it, which comes to roughly 32k generated tokens per optimizer step at its released batch shape. \method draws eight rollouts under a 4096-token budget across eight prompts, or roughly 260k tokens per step. That is about eight times more generation for the same number of updates, and it costs 9 hours of training against 3.5 on identical hardware (Appendix~\ref{app:details}). The gap is wider than those totals suggest, because only the disagreeing rollouts are distilled. At $k{=}1$ at most eight sequences per step carry gradient, against 32 for supervised OPSD. \method therefore substitutes sampling compute for annotation. That substitution is advantageous when unlabeled problems are abundant and labels are costly or unreliable, and disadvantageous when a labeled corpus already exists and generation is the bottleneck. Reducing $G$ or amortizing votes across epochs would narrow the gap, and we leave both to future work.
\begin{table}[h]
\caption{Performance comparison on math reasoning benchmarks for Qwen3 models with thinking mode. }
\label{tab:scale_th}
\centering
\small
\setlength{\tabcolsep}{5pt}
\begin{tabular}{cl|cccccc}
\toprule
 & \textbf{Method} & \textbf{AIME24} & \textbf{AIME25} & \textbf{HMMT25} & \textbf{MATH500} & \textbf{AMC23} & \textbf{Avg.} \\
\midrule
\multicolumn{8}{l}{\textit{Qwen3-4B}} \\
 & \quad Base  & \cellcolor{gradblue!12}74.17 & 64.72 & \cellcolor{gradblue!24}45.56 & 94.80 & \cellcolor{gradblue!4}95.00 & \cellcolor{gradblue!2}74.85 \\
\multirow{3}{*}{\emph{w/ GT.}} & \quad + SFT & \cellcolor{gradblue!2}73.06 & \cellcolor{gradblue!30}69.17 & 41.67 & \cellcolor{gradblue!17}95.40 & 94.38 & 74.73 \\
 & \quad + GRPO & \cellcolor{gradblue!10}73.89 & \cellcolor{gradblue!32}\textbf{69.44} & \cellcolor{gradblue!12}43.61 & \cellcolor{gradblue!19}95.45 & \cellcolor{gradblue!32}\textbf{99.38} & \cellcolor{gradblue!22}76.35 \\
 & \quad + OPSD & \cellcolor{gradblue!22}75.28 & \cellcolor{gradblue!23}68.06 & \cellcolor{gradblue!8}43.06 & \cellcolor{gradblue!12}95.20 & \cellcolor{gradblue!32}\textbf{99.38} & \cellcolor{gradblue!20}76.20 \\
\addlinespace[1.5pt]
\cdashline{2-8}[1.2pt/1.4pt]
\addlinespace[1.5pt]
\multirow{4}{*}{\emph{w/o GT.}} & \quad + TTRL & 72.78 & \cellcolor{gradblue!24}68.33 & \cellcolor{gradblue!24}45.56 & \cellcolor{gradblue!32}\textbf{95.90} & \cellcolor{gradblue!12}96.25 & \cellcolor{gradblue!14}75.76 \\
 & \quad + RENT & \cellcolor{gradblue!17}74.72 & \cellcolor{gradblue!8}65.83 & \cellcolor{gradblue!8}43.06 & \cellcolor{gradblue!13}95.25 & \cellcolor{gradblue!32}\textbf{99.38} & \cellcolor{gradblue!13}75.65 \\
 & \quad + Intuitor & \cellcolor{gradblue!32}\textbf{76.39} & \cellcolor{gradblue!24}68.33 & \cellcolor{gradblue!7}42.78 & \cellcolor{gradblue!16}95.35 & \cellcolor{gradblue!20}97.50 & \cellcolor{gradblue!18}76.07 \\
 & \quad + \method & \cellcolor{gradblue!32}\textbf{76.39} & \cellcolor{gradblue!23}68.06 & \cellcolor{gradblue!32}\textbf{46.94} & \cellcolor{gradblue!28}95.75 & \cellcolor{gradblue!24}98.12 & \cellcolor{gradblue!32}\textbf{77.05} \\
\midrule
\multicolumn{8}{l}{\textit{Qwen3-8B}} \\
 & \quad Base & 75.56 & 66.67 & \cellcolor{gradblue!9}45.00 & \cellcolor{gradblue!32}\textbf{96.35} & \cellcolor{gradblue!19}96.88 & 76.09 \\
\multirow{3}{*}{\emph{w/ GT.}} & \quad + SFT & \cellcolor{gradblue!5}76.39 & \cellcolor{gradblue!21}69.72 & 43.89 & \cellcolor{gradblue!5}95.80 & 95.00 & \cellcolor{gradblue!1}76.16 \\
 & \quad + GRPO & \cellcolor{gradblue!8}76.94 & \cellcolor{gradblue!17}69.17 & \cellcolor{gradblue!32}\textbf{47.78} & 95.70 & 95.00 & \cellcolor{gradblue!14}76.92 \\
 & \quad + OPSD & \cellcolor{gradblue!32}\textbf{80.83} & \cellcolor{gradblue!21}69.72 & \cellcolor{gradblue!23}46.67 & \cellcolor{gradblue!2}95.75 & \cellcolor{gradblue!19}96.88 & \cellcolor{gradblue!32}77.97 \\
\addlinespace[1.5pt]
\cdashline{2-8}[1.2pt/1.4pt]
\addlinespace[1.5pt]
\multirow{4}{*}{\emph{w/o GT.}} & \quad + TTRL & \cellcolor{gradblue!10}77.22 & \cellcolor{gradblue!13}68.61 & \cellcolor{gradblue!25}46.94 & \cellcolor{gradblue!2}95.75 & \cellcolor{gradblue!13}96.25 & \cellcolor{gradblue!14}76.95 \\
 & \quad + RENT & \cellcolor{gradblue!12}77.50 & \cellcolor{gradblue!24}70.28 & \cellcolor{gradblue!16}45.83 & \cellcolor{gradblue!12}95.95 & \cellcolor{gradblue!13}96.25 & \cellcolor{gradblue!18}77.16 \\
 & \quad + Intuitor & \cellcolor{gradblue!8}76.94 & \cellcolor{gradblue!24}70.28 & \cellcolor{gradblue!2}44.17 & \cellcolor{gradblue!25}96.20 & \cellcolor{gradblue!6}95.62 & \cellcolor{gradblue!9}76.64 \\
 & \quad + \method & \cellcolor{gradblue!8}76.94 & \cellcolor{gradblue!32}\textbf{71.39} & \cellcolor{gradblue!30}47.50 & \cellcolor{gradblue!15}96.00 & \cellcolor{gradblue!32}\textbf{98.12} & \cellcolor{gradblue!32}\textbf{77.99} \\
\bottomrule
\end{tabular}
\end{table}

\paragraph{Non-thinking mode.}
\label{sec:exp:scale}
Table~\ref{tab:scale_nt} reports the results in non-thinking mode. \method achieves average scores of 49.49 and 54.31 on Qwen3-4B and Qwen3-8B, improving over the corresponding base models by 8.5\% and 10.7\%. Despite using no ground-truth solutions, \method outperforms all supervised baselines, including SFT, GRPO, and OPSD, exceeding OPSD by 3.2\% and 2.3\% at the two model scales. These gains are consistent across benchmarks: \method obtains the best result on four of the five benchmarks for both models. In contrast, the label-free RL baselines improve over the base models by at most 1.5\% under the same rollout budget. This comparison highlights the advantage of using consensus-derived solutions as privileged teacher context for token-level distillation, rather than reducing them to scalar rewards.

\paragraph{Thinking mode.}
Table~\ref{tab:scale_th} presents the results in thinking mode. \method reaches 77.05 on Qwen3-4B and 77.99 on Qwen3-8B, improving over the base models by 2.2\% and 1.9\%, respectively. It matches or slightly exceeds supervised OPSD and consistently outperforms GRPO in average performance. The smaller gains relative to non-thinking mode suggest that consensus-based self-distillation is most effective when the base model is sufficiently reliable to produce informative consensus while retaining substantial room for improvement; the stronger thinking-mode models leave less such headroom.

\paragraph{Instruction-tuned models.}
Table~\ref{tab:instruct} further evaluates \method on two instruction-tuned models. \method improves Qwen3-30B-A3B-Instruct-2507 from 75.77 to 77.46 and Qwen3-4B-Instruct-2507 from 67.00 to 68.78, achieving the highest average performance on both models. It also surpasses supervised OPSD by 1.1\% and 1.7\% on average, though on Qwen3-4B-Instruct-2507 OPSD leads on HMMT25, MATH500 and AMC23. The result on Qwen3-30B-A3B-Instruct-2507 demonstrates that the same training recipe transfers effectively to a substantially larger mixture-of-experts model without model-specific hyperparameter tuning.

\begin{table}[t]
\caption{Performance comparison on math reasoning benchmarks for two instruction-tuned models, Qwen3-30B-A3B-Instruct-2507 and Qwen3-4B-Instruct-2507.}
\label{tab:instruct}
\centering
\small
\setlength{\tabcolsep}{2.6pt}
\begin{tabular}{llcccccc}
\toprule
\bf{Model} & \bf{Method} & \bf{AIME24} & \bf{AIME25} & \bf{HMMT25} & \bf{MATH500} & \bf{AMC23} & \bf{Avg.} \\
\midrule
\multirow{3}{*}{\shortstack[l]{Qwen3-30B-A3B\\-Instruct-2507}} & Base & \cellcolor{gradblue!32}\textbf{80.00} & \cellcolor{gradblue!16}63.33 & 43.33 & 96.33 & 95.83 & 75.77 \\
 & OPSD & \cellcolor{gradblue!24}78.89 & 61.11 & \cellcolor{gradblue!21}47.78 & \cellcolor{gradblue!32}\textbf{97.20} & \cellcolor{gradblue!8}96.67 & \cellcolor{gradblue!11}76.33 \\
 & \method & 75.56 & \cellcolor{gradblue!32}\textbf{65.56} & \cellcolor{gradblue!32}\textbf{50.00} & \cellcolor{gradblue!25}97.00 & \cellcolor{gradblue!32}\textbf{99.17} & \cellcolor{gradblue!32}\textbf{77.46} \\
\cmidrule{2-8}
\multirow{3}{*}{\shortstack[l]{Qwen3-4B\\-Instruct-2507}} & Base & \cellcolor{gradblue!21}66.67 & \cellcolor{gradblue!6}53.33 & 27.78 & 93.87 & 93.33 & 67.00 \\
 & OPSD & 62.22 & 52.22 & \cellcolor{gradblue!32}\textbf{31.11} & \cellcolor{gradblue!32}\textbf{94.93} & \cellcolor{gradblue!32}\textbf{95.00} & \cellcolor{gradblue!2}67.10\\
 & \method & \cellcolor{gradblue!32}\textbf{68.89} & \cellcolor{gradblue!32}\textbf{57.78} & \cellcolor{gradblue!11}28.89 & \cellcolor{gradblue!10}94.20& \cellcolor{gradblue!16}94.17 & \cellcolor{gradblue!32}\textbf{68.78} \\
\bottomrule
\end{tabular}
\end{table}

\begin{figure}[t]
\centering
\includegraphics[width=1.0\linewidth]{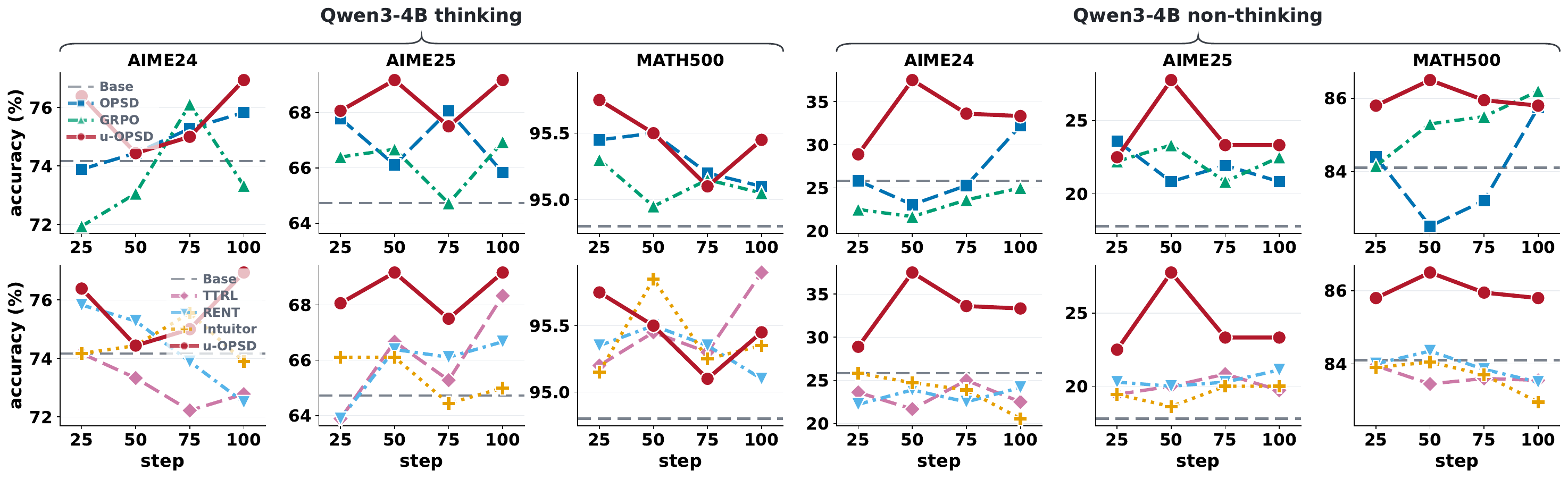}
\caption{Training curves of the Qwen3-4B thinking (left) and non-thinking mode (right) runs on AIME24, AIME25 and MATH500. \textbf{Top:} \method against the supervised methods. \textbf{Bottom:} \method against the label-free methods.}
\label{fig:curves}
\end{figure}

\paragraph{Training curves.}
Figure~\ref{fig:curves} plots the accuracy over checkpoints of the Qwen3-4B runs on AIME24, AIME25 and MATH500 in both modes. The two modes separate immediately. In thinking mode every method stays within about 2\% of the base model and the ordering changes from checkpoint to checkpoint, with the whole MATH500 panel spanning less than 1\%. The label-free baselines track the base model closely, as do \method and OPSD. In non-thinking mode \method is above every other method at every checkpoint on both AIME benchmarks, against the supervised and the label-free baselines alike; on MATH500 the other methods catch up to the same level by step 100. 

\subsection{Ablation and analysis}
\label{sec:exp:ablations}

\paragraph{Pseudo-label quality.}
\label{sec:exp:probe}

On 64 training prompts under the full training configuration ($G{=}8$, 4096-token budget, $\tau{=}0.5$), 96.3\% of rollouts yield a parsable boxed answer, 94.0\% of prompts clear the self-consistency threshold and receive a pseudo-label, and 86.7\% of those pseudo-labels match the gold answer from data source (used in supervised methods). Fewer than 10\% of valid rollouts disagree with their prompt's vote, so the distillation set is small and concentrated.

\begin{figure}[t]
\centering
\begin{tikzpicture}
\begin{groupplot}[
  group style={group size=3 by 1, horizontal sep=0.75cm},
  width=0.380\linewidth, height=0.34\linewidth,
  tick label style={font=\small, color=cInk}, label style={font=\small, color=cInk},
  axis line style={color=cInk, line width=0.8pt},
  grid=both, grid style={color=cInk!45, line width=0.8pt, dotted},
  tick align=outside, tick pos=left, tick style={color=cInk, line width=0.7pt},
]
% ---- threshold tau
\nextgroupplot[
  xlabel={self-consistency threshold $\tau$}, ylabel={Averaged Accuracy (\%)},
  xmin=0.24, xmax=0.96, xtick={0.3,0.5,0.7,0.9},
  ymin=41, ymax=71,
  legend style={at={(0.97,0.97)}, anchor=north east, font=\scriptsize, draw=none, fill=none, row sep=0pt, inner sep=1pt},
  every axis plot/.append style={line width=1.7pt, mark size=2.3pt, mark options={solid}},
]
\addplot[mark=square*, color=cRedS] coordinates {(0.3,58.59)(0.5,57.10)(0.7,48.18)(0.9,44.40)};
\addlegendentry{best ckpt}
\addplot[mark=*, color=cBlueS] coordinates {(0.3,56.08)(0.5,53.93)(0.7,45.96)(0.9,44.40)};
\addlegendentry{step 150}
\draw[dashed, color=cBase, line width=0.8pt] (axis cs:0.5,41) -- (axis cs:0.5,66);
\node[font=\scriptsize, color=cBase, anchor=south west] at (axis cs:0.51,41.4) {default};
% ---- rollouts per prompt G
\nextgroupplot[
  ybar=0pt, bar width=9pt, bar shift auto,
  xlabel={rollouts per prompt $G$},
  symbolic x coords={4,8,12,16}, xtick=data,
  ymin=50, ymax=70, enlarge x limits=0.2,
  legend style={at={(0.03,0.97)}, anchor=north west, font=\scriptsize, draw=none, fill=none, row sep=0pt, inner sep=1pt},
  legend image code/.code={\draw[#1,draw=none] (0cm,-0.08cm) rectangle (0.26cm,0.08cm);},
  every axis plot/.append style={draw=none},
]
\addplot[fill=cBlueS, fill opacity=0.8, draw=none] coordinates {(4,54.73)(8,53.93)(12,58.45)(16,58.59)};
\addlegendentry{step 150}
\addplot[fill=cRedS, fill opacity=0.8, draw=none] coordinates {(4,56.99)(8,57.10)(12,61.79)(16,59.37)};
\addlegendentry{best ckpt}
% ---- teacher update
\nextgroupplot[
  ybar=0pt, bar width=9pt, bar shift auto,
  xlabel={teacher update},
  symbolic x coords={fixed,0.99,0.995,0.999}, xtick=data,
  xticklabels={fixed,{EMA\\0.99},{EMA\\0.995},{EMA\\0.999}},
  x tick label style={font=\scriptsize, color=cInk, align=center},
  ymin=52, ymax=61, enlarge x limits=0.2,
  every axis plot/.append style={draw=none},
]
\addplot[fill=cBlueS, fill opacity=0.8, draw=none] coordinates {(fixed,53.93)(0.99,57.89)(0.995,58.05)(0.999,54.90)};
\addplot[fill=cRedS, fill opacity=0.8, draw=none] coordinates {(fixed,57.10)(0.99,58.88)(0.995,59.47)(0.999,58.82)};
\end{groupplot}
\end{tikzpicture}
\caption{Configuration ablations, on Qwen3-8B non-thinking. The vertical axis is accuracy averaged over the five evaluation benchmarks. \textbf{Left:} self-consistency threshold $\tau$, the fraction of valid rollouts that must agree for a prompt to receive a pseudo-label. \textbf{Middle:} rollouts per prompt $G$, at $\tau{=}0.5$. \textbf{Right:} how the teacher is updated; ``fixed'' freezes it at the initial policy and is our default, while the EMA rows let it track the student at the given decay.}
\label{fig:config}
\end{figure}
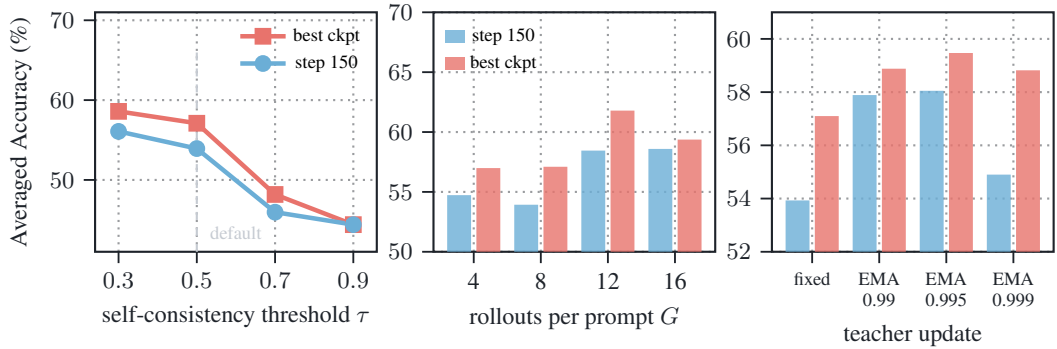

\paragraph{Self-consistency threshold.}
The threshold $\tau$ decides which prompts are allowed to supervise. Figure~\ref{fig:config} (left) sweeps it and the ranking is monotone: the loosest threshold is best, at 58.59 for $\tau{=}0.3$ against 57.10 for the default and 44.40 at $\tau{=}0.9$, a 14.2\% spread; the step-150 series in the figure is ordered the same way. The default $\tau{=}0.5$ is therefore not the best setting in this sweep. %Low-agreement prompts still carry usable signal despite their unreliable pseudo-labels, or the threshold mostly removes hard prompts and with them the part of the distribution where headroom remains.%

\paragraph{Rollouts per prompt.}
$G$, number of rollouts per prompt, controls both the resolution of the vote and the pool from which disagreeing rollouts are drawn, and it is the dominant cost term, since the sampling cost grows linearly in $G$. Figure~\ref{fig:config} (middle) shows the return is real but soon saturating: $G{=}4$ and $G{=}8$ are indistinguishable at 56.99 and 57.10 at the best checkpoint, $G{=}12$ gains 4.7\% over the default, and $G{=}16$ gives back half of that. We keep $G{=}8$ elsewhere because it is OPSD's own generation budget and makes the comparison in Tables~\ref{tab:scale_nt} and~\ref{tab:scale_th} cost-matched, but $G{=}12$ is the better operating point when generation is not the constraint.

\paragraph{Teacher update.}
OPSD \citep{zhao2026self} freezes the teacher at the initial policy, which under LoRA means evaluating the base weights with the adapter disabled. We report the frozen teacher elsewhere because it is OPSD's setting and keeps the comparison in Tables~\ref{tab:scale_nt} and~\ref{tab:scale_th} matched. Yet, Figure~\ref{fig:config} (right) shows that the EMA settings offer more room for improvement: the best, at decay 0.995, gains 2.4\% over the frozen default at the best checkpoint and 4.1\% at step 150, and the two faster decays land within 0.2\% of each other at step 150. 

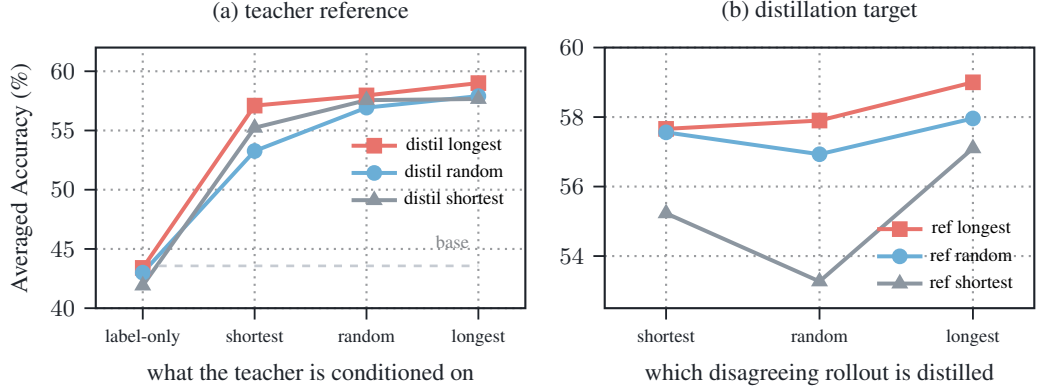
\begin{figure}[t]\centering
\begin{tikzpicture}
\begin{groupplot}[
  group style={group size=2 by 1, horizontal sep=1.05cm},
  width=0.52\linewidth, height=0.36\linewidth,
  tick label style={font=\small, color=cInk}, label style={font=\small, color=cInk},
  title style={font=\small, color=cInk},
  axis line style={color=cInk, line width=0.8pt},
  grid=both, grid style={color=cInk!45, line width=0.8pt, dotted},
  tick align=outside, tick pos=left, tick style={color=cInk, line width=0.7pt},
  every axis plot/.append style={line width=1.5pt, mark size=2.2pt, mark options={solid}},
  legend style={font=\scriptsize, draw=none, fill=none, row sep=0pt, inner sep=1pt},
]
% ---- (a) teacher reference
\nextgroupplot[
  title={(a) teacher reference},
  xlabel={what the teacher is conditioned on}, ylabel={Averaged Accuracy (\%)},
  symbolic x coords={label-only,shortest,random,longest}, xtick=data,
  x tick label style={font=\scriptsize, color=cInk},
  ymin=40, ymax=62, enlarge x limits=0.14,
  legend style={at={(0.97,0.35)}, anchor=south east},
]
\addplot[mark=square*, color=cRedS]  coordinates {(label-only,43.40)(shortest,57.10)(random,57.96)(longest,59.00)};
\addlegendentry{distil longest}
\addplot[mark=*, color=cBlueS]       coordinates {(label-only,43.00)(shortest,53.27)(random,56.93)(longest,57.90)};
\addlegendentry{distil random}
\addplot[mark=triangle*, color=cGreyS] coordinates {(label-only,41.91)(shortest,55.23)(random,57.56)(longest,57.66)};
\addlegendentry{distil shortest}
\draw[dashed, color=cBase, line width=1pt] (axis cs:label-only,43.57) -- (axis cs:longest,43.57);
\node[font=\scriptsize, color=cInk!55, anchor=south east] at (axis cs:longest,44.3) {base};
% ---- (b) distillation target
\nextgroupplot[
  title={(b) distillation target},
  xlabel={which disagreeing rollout is distilled},
  symbolic x coords={shortest,random,longest}, xtick=data,
  x tick label style={font=\scriptsize, color=cInk},
  ymin=52.5, ymax=60, enlarge x limits=0.20,
  legend style={at={(0.97,0.03)}, anchor=south east},
]
\addplot[mark=square*, color=cRedS]  coordinates {(shortest,57.66)(random,57.90)(longest,59.00)};
\addlegendentry{ref longest}
\addplot[mark=*, color=cBlueS]       coordinates {(shortest,57.56)(random,56.93)(longest,57.96)};
\addlegendentry{ref random}
\addplot[mark=triangle*, color=cGreyS] coordinates {(shortest,55.23)(random,53.27)(longest,57.10)};
\addlegendentry{ref shortest}
\end{groupplot}
\end{tikzpicture}
\caption{\textbf{Teacher reference against distillation target}, on Qwen3-8B non-thinking, $G{=}8$, $\tau{=}0.5$, $k{=}1$; each point is the five-benchmark average. \emph{(a)} What the teacher is conditioned on. \emph{(b)} Which disagreeing rollout is distilled.}
\label{fig:refgrid}
\end{figure}

\paragraph{Teacher reference and distillation target.}
Figure~\ref{fig:refgrid} crosses which agreeing rollout the teacher is conditioned on against which disagreeing rollout is distilled. Both axes favour the longest rollout, though choosing the shortest or a random one also yields competitive performance: seven of the nine full-reference settings lie within 2.1\% of the best on averaged benchmark accuracy, with random the weaker choice as a distillation target, at a column mean of 56.03 against 58.02 for longest. The best setting, at 59.00, conditions the teacher on the longest agreeing rollout and distils the longest disagreeing one. Yet, as (a) shows, further stripping the reference to the boxed pseudo-label costs 10.3\% to 15.8\% and puts all three label-only settings below the base model's 43.57. Effective distillation requires the teacher to condition on the longer, full reasoning trace rather than only the final answer.

\begin{table}[t]
\caption{Comparison of divergence computation strategy: Full vocabulary is logit distillation over every token~\citep{agarwal2024gkd}; sampled token evaluates the two policies only at the token the student drew~\citep{lu2025onpolicydistillation}; top-$k$ rows truncate the teacher to its $k$ largest entries. We report on Qwen3-8B non-thinking at the best checkpoint.}
\label{tab:vocab}
\centering
\small
\setlength{\tabcolsep}{4pt}
\begin{tabular}{l|ccccccc}
\toprule
\multicolumn{2}{l}{Variant} & AIME24 & AIME25 & HMMT25 & MATH500 & AMC23 & Avg. \\
\midrule
\multicolumn{2}{l}{Student token} & 29.44 & 19.72 & 12.78 & 84.70 & 70.62 & 43.45 \\
\multicolumn{2}{l}{Full-vocabulary} & \cellcolor{gradblue!32}\textbf{53.89} & \cellcolor{gradblue!11}37.50 & \cellcolor{gradblue!4}20.00 & \cellcolor{gradblue!4}89.75 & \cellcolor{gradblue!8}84.38 & \cellcolor{gradblue!13}57.10 \\
\addlinespace[1.5pt]
\cdashline{1-8}[1.2pt/1.4pt]
\addlinespace[1.5pt]
\multirow{4}{*}{\quad top-$k$} & 20 & \cellcolor{gradblue!6}49.17 & \cellcolor{gradblue!12}37.78 & 19.44 & \cellcolor{gradblue!11}90.20 & \cellcolor{gradblue!32}\textbf{88.12} & \cellcolor{gradblue!11}56.94 \\
 & 50 & 48.06 & 34.44 & \cellcolor{gradblue!32}\textbf{24.17} & 89.45 & 83.12 & 55.85 \\
 & 100 & \cellcolor{gradblue!32}\textbf{53.89} & \cellcolor{gradblue!32}\textbf{43.33} & \cellcolor{gradblue!8}20.56 & \cellcolor{gradblue!32}\textbf{91.65} & \cellcolor{gradblue!16}85.62 & \cellcolor{gradblue!32}\textbf{59.01} \\
 & 200 & \cellcolor{gradblue!20}51.67 & \cellcolor{gradblue!20}40.00 & \cellcolor{gradblue!21}22.50 & \cellcolor{gradblue!10}90.15 & \cellcolor{gradblue!20}86.25 & \cellcolor{gradblue!23}58.11 \\
\bottomrule
\end{tabular}
\end{table}

\begin{table}[t]
\caption{Divergence metrics $D_{\beta}$ under \method, on Qwen3-8B non-thinking, longest-1, $G{=}8$, $\tau{=}0.5$.}
\label{tab:beta}
\centering
\small
\setlength{\tabcolsep}{3pt}
\begin{tabular}{lcccccc}
\toprule
Objective & AIME24 & AIME25 & HMMT25 & MATH500 & AMC23 & Avg. \\
\midrule
Base & 27.50 & \cellcolor{gradblue!3}23.33 & \cellcolor{gradblue!6}13.61 & \cellcolor{gradblue!2}84.05 & 69.38 & \cellcolor{gradblue!1}43.57 \\
Forward KL $\mathrm{KL}(\pi_T \parallel \pi_S)$, $\beta{=}0$ (default) & \cellcolor{gradblue!32}\textbf{53.89} & \cellcolor{gradblue!32}\textbf{37.50} & \cellcolor{gradblue!32}\textbf{20.00} & \cellcolor{gradblue!32}\textbf{89.75} & \cellcolor{gradblue!32}\textbf{84.38} & \cellcolor{gradblue!32}\textbf{57.10} \\
Reverse KL $\mathrm{KL}(\pi_S \parallel \pi_T)$, $\beta{=}1$ & \multicolumn{6}{c}{\textit{training diverges (not scored)}} \\
JSD ($\beta{=}0.5$) & \cellcolor{gradblue!2}29.44 & 21.94 & 12.22 & 83.70 & 69.38 & 43.34 \\
\bottomrule
\end{tabular}
\end{table}

\paragraph{Divergence computation strategy.}
Full-vocabulary logit distillation~\citep{agarwal2024gkd} evaluates $D_{\beta}$ over the entire vocabulary at each position, giving a proper token-level divergence between the two policies. Sampled-token distillation~\citep{lu2025onpolicydistillation} evaluates teacher and student log-probabilities only at the token the student sampled, and uses that term as a scalar advantage inside a policy-gradient objective. OPSD \citep{zhao2026self} reports the full-distribution objective to be the stronger of the two under gold supervision, and because \method changes the teacher's conditioning context rather than its objective, the open question is whether that preference survives once the privileged context is a pseudo-solution. Table~\ref{tab:vocab} shows that it does, and by a wider margin: the full-distribution objective leads by 17.8\% on AIME25 and 7.2\% on HMMT25, against 2.0\% and 2.7\% under gold supervision. The student token-only variant is 13.7\% below the default and 15.6\% below the best setting, so it is not competitive under pseudo-label supervision. Yet these tend to be the extremes of one axis, so we further test whether an intermediate solution suffices: restricting the divergence to the teacher's $k$ largest entries, from $k{=}20$ upward. The best score on each of the five benchmarks falls mostly under top-100, and the full vocabulary exceeds none of them on average, tying top-100 on AIME24 and trailing elsewhere. The top-$k$ truncation is therefore a practical optimization that reduces the cost of computing the divergence.

\paragraph{Divergence Metrics.}
OPSD uses forward KL with $\beta{=}0$ and table~\ref{tab:beta} shows the preference towards forward KL is stronger under \method. Using symmetric Jensen--Shannon divergence (JSD) costs 13.8\%, which puts it level with the untrained model, and reverse KL does not produce a converged model. The collapse under reverse KL is gradual and one-directional: generations average 2.7k characters at step 25, 12k at step 50, 76k at step 75, and 99k --- the token ceiling --- from step 125 on, while the fraction of rollouts that yield a parsable boxed answer falls from 99\% to 33\%. Inspection of the completions shows a loss of termination rather than cohesive reasoning (e.g., a phrase, a \LaTeX{} command pair, or a nesting bracket repeated until the token budget runs out).

\section{Limitations}
\label{sec:limitations}

\textbf{Scope.} Our experiments cover one model family (Qwen3, at 4B and 8B and in both reasoning modes) and one domain (competition mathematics) with automatically checkable final answers. The voting mechanism requires an extractable, canonicalizable answer; extending \method to open-ended generation would require replacing exact-match voting with a softer consensus.

\textbf{The size of the advantage is regime-dependent.} The gain over supervised OPSD is large in non-thinking mode and less in thinking mode. \method exceeds OPSD by 3.2\% and 2.3\% at 4B and 8B in Table~\ref{tab:scale_nt}, and by 0.9\% and 0.02\% in Table~\ref{tab:scale_th}; the improvement over the base model shrinks in the same way, from 8.5\% and 10.7\% to 2.2\% and 1.9\%. Thinking-mode base accuracy is already 74.9 and 76.1, and thinking-mode rollouts are far longer, so a 150-step budget covers many fewer completed votes per token spent. Readers should therefore treat consensus as a \emph{stronger replacement} for gold solutions in the non-thinking regime we measured and as a \emph{match} for them in the thinking regime, rather than as a uniform improvement on on-policy self-distillation.

\textbf{Dependence on base-model competence.} Majority-vote supervision reproduces whichever answer the base model already produces most often, so it is bounded by that answer's accuracy. Our probe measured 13.3\% wrong pseudo-labels in-domain, and we did not measure training dynamics under deliberately corrupted votes. Mechanisms for detecting or down-weighting low-quality consensus (e.g., by vote margin) are natural next steps.

\textbf{Variance and repeats.} We mitigated variance by evaluating every checkpoint with 12 samples spanning five benchmarks, but seed-replicated error bars for every training setup are pending and will be added in a revision.

\section{Conclusion}

We showed that the ground-truth solution in on-policy self-distillation can be replaced by the model's own majority-vote consensus: the agreeing rollout serves as the teacher's reference, and distillation is applied only to self-inconsistent rollouts. With the OPSD recipe otherwise untouched, this label-free variant outperforms its supervised counterpart on the five-benchmark average, and a grid over two model sizes and both reasoning modes shows the advantage reproduces at both scales in non-thinking mode and a tie in thinking mode, where the base model is already strong enough that little headroom remains. The result suggests that, in the regime of on-policy self-distillation---a base model competent enough to vote well and fallible enough to have headroom---the binding constraint is not access to gold solutions but the machinery for surfacing and correcting the model's own inconsistencies, a machinery that requires no supervision.

\bibliographystyle{plainnat}
\bibliography{references}

\newpage

\end{document}